\documentclass[runningheads]{llncs}
\usepackage[T1]{fontenc}
\usepackage{graphicx}
\usepackage{color}
\usepackage{xspace}
\usepackage{graphicx}
\usepackage{amsmath}
\usepackage{amssymb}
\usepackage{multirow}

\makeatletter
\DeclareRobustCommand\onedot{\futurelet\@let@token\@onedot}
\def\@onedot{\ifx\@let@token.\else.\null\fi\xspace}

\def\eg{\emph{e.g}\onedot} 
\def\ie{\emph{i.e}\onedot}

\def\etal{\emph{et al}\onedot}
\makeatother

\begin{document}
\title{CLAP: Concave Linear APproximation for Quadratic Graph Matching}

\titlerunning{CLAP}
%
\author{Yongqing Liang\inst{1}\orcidID{0000-0002-7282-0476} \and
Huijun Han\inst{1}\orcidID{0009-0009-1360-4886} \and
Xin Li\inst{1}\orcidID{0000-0002-0144-9489}}
\authorrunning{Liang et al.}
%
\institute{Texas A\&M University, College Station TX 77840, USA \\
\email{\{lyq, hazelhan, xinli\}@tamu.edu}
}

\maketitle              
\begin{abstract}

Solving point-wise feature correspondence in visual data is a fundamental problem in computer vision. 
A powerful model that addresses this challenge is to formulate it as graph matching, which entails solving a Quadratic Assignment Problem (QAP) with node-wise and edge-wise constraints. 
However, solving such a QAP can be both expensive and difficult due to numerous local extreme points. 
In this work, we introduce a novel linear model and solver designed to accelerate the computation of graph matching. 
Specifically, we employ a positive semi-definite matrix approximation to establish the structural attribute constraint.
We then transform the original QAP into a linear model that is concave for maximization. 
This model can subsequently be solved using the Sinkhorn optimal transport algorithm, known for its enhanced efficiency and numerical stability compared to existing approaches. 
Experimental results on the widely used benchmark PascalVOC showcase that our algorithm achieves state-of-the-art performance with significantly improved efficiency. 
We plan to release our code for public access.

\end{abstract}

\begin{figure}[t]
    \begin{center}
        \includegraphics[width=0.7\linewidth]{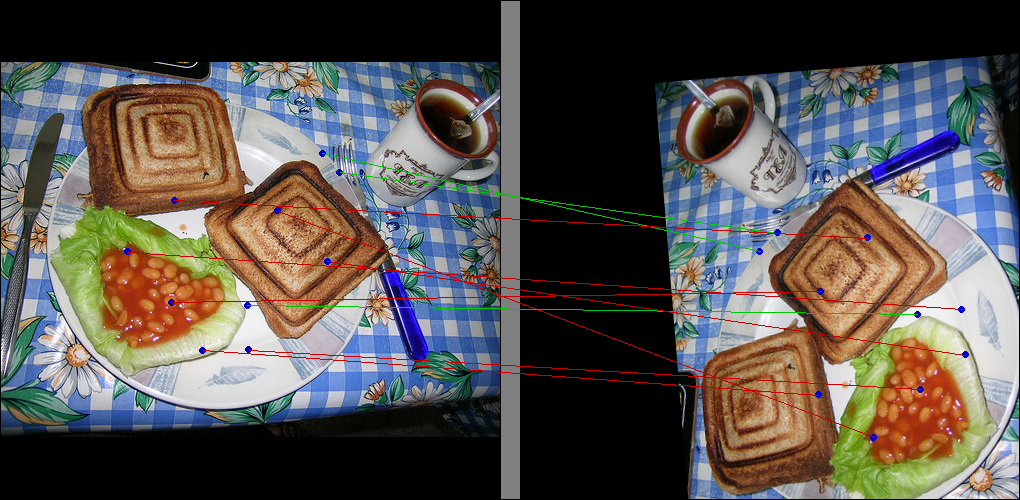}\\
        (a) Graph matchings of unary matching CIE-H~\cite{yu2019learningCIEH}.\\
        \includegraphics[width=0.7\linewidth]{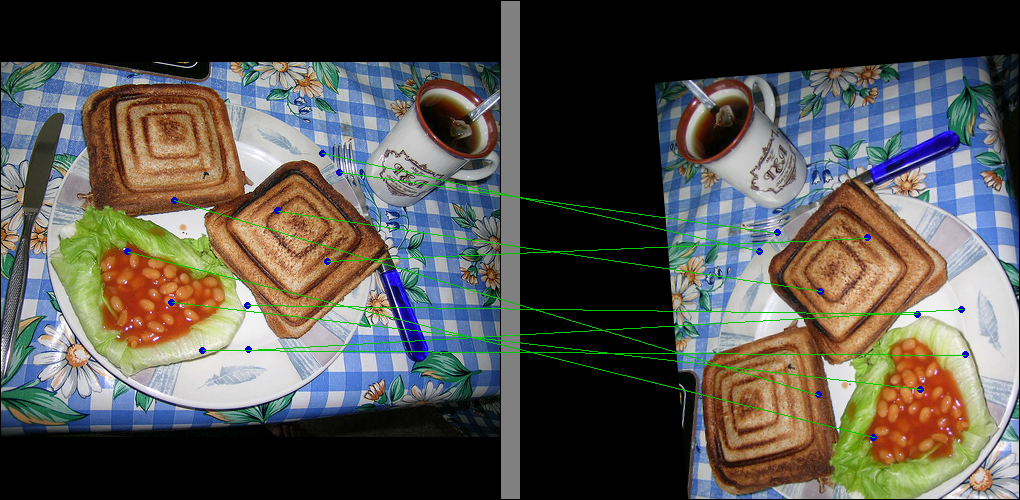}\\
        (b) Graph matchings of our pairwise matching model.\\
    \end{center}
    \caption{Graph matching results by unary matching and pairwise matching. The image undergoes an affine transformation. \textcolor{green}{Green}/\textcolor{red}{red} lines indicate correct/wrong matchings.}
    \label{fig:synth-unary-pairwise}
\end{figure}

\section{Introduction}
\label{sec:intro}

Graph is a natural structure to encode real-world data. 
Graph matching is to find node-to-node correspondences between two given graphs. 
Hence, as a general and powerful tool to discover correlation or detect similar structural pattern between graphs, graph matching has been widely used in many computer vision tasks including keypoints matching~\cite{sun2021loftr}, multi-object tracking~\cite{he2021learnable}, and scene flow estimation~\cite{puy2020flot}. 
A graph matching pipeline starts from extracting keypoints and their descriptors. 
Then, each image/frame/object is modeled using a graph, whose nodes correspond to keypoints or regions and are associated with their descriptors (i.e., \emph{attributes}) and edges encode relationship between these nodes. 
And then, the matching is solved through certain optimization procedure that finds a node-to-node map that minimizes content/structural matching cost. 
Recent graph matching methods can be divided into two categories~\cite{wang2020combinatorialPCA}, \emph{Unary Matching} and \emph{Pairwise Matching}. 

\emph{Unary Matching} methods~\cite{wang2019learningIPCA,sun2021loftr,sarlin2020superglue} formulate graph matching as node matching. 
Deep neural networks are often used to learn discriminating descriptors on keypoints (nodes); and these descriptors can be designed to encode both local context surrounding nodes and inter-node relationship. 
Then, subsequent matching can be computed using similarity between these descriptors (namely, \emph{pointwise affinity}). 
Unary matching methods can run fast and achieve competitive results on common benchmarks. 
However, as pointed out in recent studies~\cite{gao_deep_2021}, when keypoints share similar descriptors but different structural contexts, unary methods often fail to produce reliable matching. Fig.~\ref{fig:synth-unary-pairwise}~(a) also illustrates such an example: these two images undergo an affine transformation, a unary matching method such as CIE-H~\cite{yu2019learningCIEH} matches two graphs by node attributes that are extracted from a deep neural network. 
Unary matching methods often fail to find the correct correspondence when node attributes become indistinctive (because (1) the point's neighboring textures are not unique, and (2) transformations involving significant rotations make deep features less reliable). 

\emph{Pairwise Matching} methods~\cite{zaslavskiy2008path,wang_functional_2020,zhang2019kergm,gao_deep_2021} construct the matching model using not only \emph{pointwise affinity} information but also \emph{pairwise structure} constraints and formulate graph matching as a Koopmans-Beckman Quadratic Assignment Problem (KB-QAP)~\cite{koopmans1957assignment}.
People integrate local features from adjacent nodes to compose \emph{edge attributes} to encode the inter-node structure information. 
Then these methods develop pairwise structure constraint to penalize edge attributes discrepancy according to node correspondence.

Pairwise structure constraints make the matching more robust against global (camera-wise) and local (object-related) geometric transformations~\cite{gao_deep_2021,wang_functional_2020}. 
As shown in Fig.~\ref{fig:synth-unary-pairwise}~(b), with the help of structure constraint, a pairwise matching model (\eg, our model) could find correct correspondence. 
However, The objective function of KB-QAP is nonconvex-nonconcave because the Hessian matrix of its quadratic term is indefinite. 
The solution of the KB-QAP has many local maximums or minimums. 
These local extremes make the solver sensitive to initial poses; and the slow convergence makes the solver computationally expensive. 
This remains as the bottleneck of pairwise matching models, and limit their applications on real-time tasks such as multi-object tracking and others. 

In this paper, we propose a linear model, named \emph{CLAP}, to convert the pairwise graph matching to a concave maximization problem. 
We follow the objective function of KB-QAP but formulate the pairwise structure constraint into a linear model under L1 norm. 
To build such a linear model, we leverage the decomposition property of positive semi-definite matrix.
We analyze the widely-used edge attributes (\eg, Euclidean distance~\cite{wang_functional_2020} and inner-product distance~\cite{sun2021loftr,wang2019learningIPCA}), and convert them to be positive semi-definite and construct a linear structure constraint. 

We showed that our new objective function is concave, whose maximization is easy and results in a global maximum. 
Using the Sinkhorn algorithm~\cite{cuturi_sinkhorn_2013}, 
our CLAP model can be efficiently solved by the Lagrangian multiplier method.
Experiments showed that our method achieves similar accuracy with other state-of-the-art methods but runs significantly faster.

\section{Related Work}

\subsection{Unary Matching}

Unary matching based methods~\cite{yu2019learningCIEH,wang2019learningIPCA,wang2020combinatorialPCA,detone2018superpoint,sarlin2020superglue,sun2021loftr} formulates the graph matching as node matching. 
They first extract local features for each node as node attributes, then iteratively enhance the node attributes by various feature refinement modules, such as GNN~\cite{sarlin2020superglue,wang2019learningIPCA,yu2019learningCIEH} and Transformer~\cite{sun2021loftr}.
The assumption is that graph structure can be sufficiently encoded into node attributes. Therefore they can just use the node-to-node affinity matrix by inner-product or metric learning~\cite{wang2020combinatorialPCA,wang2019learningIPCA,yu2019learningCIEH} to solve graph matching. 
With the node-to-node affinity matrix, the matching can be computed by using nearest neighbor search~\cite{yi2016lift,Shen19CVPR}, dual-softmax~\cite{sun2021loftr}, or optimal transport~\cite{sarlin2020superglue,wang2020combinatorialPCA,wang2019learningIPCA}.

Despite their simplicity, the unary matching models have two limitations: 
(1) Feature embedding modules are often adopted to enhance the node attributes/descriptors, but they also slow down the overall running time.
(2) More importantly, the assumption that node attributes are distinctive enough to encode graph structure sometimes doesn't hold.
For example, when the region of interest in the images do not contain rich texture information, their node attributes can be indistinguishable and ambiguous. Without the effectively modeling structure information, unary matching models often fail to produce reliable matching results.

\subsection{Pairwise Matching}
Pairwise matching methods~\cite{zanfir2018deepGMN,wang_functional_2020,gao_deep_2021} formulate the graph matching as quadratic assignment problems (QAP)~\cite{loiola2007survey}, such as Lawler QAP~\cite{lawler1963quadratic} and Koopmans-Beckman QAP (KB-QAP)~\cite{koopmans1957assignment}. 
The objective function uses node and edge similarity constraints to build affinity matrix.
When the affinity matrix of Lawler's QAP can be decomposed by inverse Kronecker product, KB-QAP is a special case of Lawler's QAP with much lower space complexity~\cite{loiola2007survey,wang_functional_2020}.
Since this condition is true in general cases, recent papers chose KB-QAP as their objective functions.
Because the discrete QAP is NP-complete~\cite{johnson1979computers}, researchers relax the feasible field into a continuous domain to find approximate solutions in polynomial time.

The bottlenecks in developing effective QAP solvers are on the problem of many local extreme points and slow convergence because the Hessian matrix of the QAP objective function is often nonconvex-nonconcave.
Various approximate algorithms have been proposed to solve the QAP function.
Umeyama~\cite{umeyama1988eigendecomposition} used the absolute values of eigenvectors of the edge attributes to construct the structure constraint, but such approximation changes the physical meaning of the edge attributes and leads to large matching errors~\cite{zaslavskiy2008path}.
Lu~\etal~\cite{lu2016fast} proposed a fast projected fixed point scheme.
FGM~\cite{zhou2012factorizedFGM} factorized the affinity matrix of Lawler's QAP into small matrices.
PATH~\cite{zaslavskiy2008path}, Gao~\etal~\cite{gao_deep_2021} and Wang~\etal~\cite{wang_functional_2020} used Frank-Wolfe method~\cite{jaggi2015global} to obtain an approximate solution.
AFAT~\cite{wang2023deep} suggested the slow convergence rate could be addressed by partial graph matching. Thus AFAT ruled out the unpaired outliers within the detected keypoints using attention-fused prediction for the number of reliable inliers in a data-driven manner. 
The noisy nature of data contaminates the robostness of deep neural networks for graph matching. As a solution, momentum distillation is exploited by COMMON~\cite{lin2023graph} to emphasize the graph consistency by gradually lowering the supervision from the ground truth correspondence.

Existing methods are generally time consuming to find the optimal correspondences because they often require dozen iterations to convergence.
The long computation time limits the application of pairwise matching methods for real-time tasks such as multi-object tracking.
Our method belongs to pairwise matching and is based on KB-QAP objective function.

\section{Algorithm}


\subsection{Problem Definition}
\label{sec:baseline}

Graph matching solves node-to-node correspondence between two given graphs $G_A = \{V_A, E_A\}$ and $G_B = \{V_B, E_B\}$, where $V$ and $E$ denote the node and edge sets.
The node sets $V_A$ and $V_B$ contain $n$ and $m$ feature keypoints extracted from $I_A$ and $I_B$ respectively. 

The similarity between two descriptors can be measured in the feature space and represented as $U \in \mathbb{R}^{n\times m}$.
Here $U_{ij}$ represents node similarity between $(v_A)_i \in V_A$ and $(v_B)_j \in V_B$.

The edge set $E$ encodes spatial (either in the positional space or feature space) correlation between two nodes in a graph. The definition of edge attributes depends on the graph matching models, \eg, Lawler QAP~\cite{lawler1963quadratic,LCSGMwang2020learning} and Koopmans-Beckman QAP (KB-QAP)~\cite{koopmans1957assignment}. 
We adopt the widely used KB-QAP model. 
The KB-QAP model uses adjacency weight matrices $D_A\in \mathbb{R}^{n\times n}$ (and $D_B\in \mathbb{R}^{m\times m}$) to store the structure information of graph $G_A$ (and $G_B$). 

The graph matching problem is to find an \emph{optimal node-to-node assignment} $P\in\{0, 1\}^{n\times m}$, where $P_{ij} = 1$ indicates that nodes $(v_A)_i \in V_A$ and $(v_B)_j \in V_B$ are matched and $P_{ij} = 0$ otherwise.
Following the graph matching setting, the feasible field of $\mathcal{P}$ is 
\begin{equation}
    \mathcal{P} \triangleq \big\{P\in \{0,1\}^{n\times m};~P\mathbf{1_m}=\mathbf{1_n},~P^T\mathbf{1_n} \le \mathbf{1_m} \big\},
\end{equation}
where $\mathbf{1_m}$ is a vector of $m$ ones.
The KB-QAP formulates graph matching as maximizing the sum of the node and the edge similarities,

\begin{equation}
    \max_{P\in\mathcal{P}} \left( \sum_{i,j}P_{ij}U_{ij} - \lambda ||D_A - PD_BP^T||_F^2  \right),
    \label{eqn:loss1}
\end{equation}
where $\lambda>0$ is a balancing weight, and $||\cdot||_F^2$ is the square of the Frobenius norm.
The columns of $P\in\mathcal{P}$ are orthogonal. 
Rewriting the second term of Eq.~(\ref{eqn:loss1}) with the trace of that matrix, we have
\begin{equation*}
    \begin{split}
        &-||D_A - PD_BP^T||_F^2 = -tr((D_A - PD_BP^T)^T(D_A - PD_BP^T))\\
        &= 2 tr(P^TD_A^TPD_B) -tr(D_A^TD_A) - tr(D_B^TD_B).
    \end{split}
    \label{eqn:quad1}
\end{equation*}
After removing the constant items, maximizing $-||D_A - PD_BP^T||_F^2$ is equivalent to maximizing $tr(P^TD_A^TPD_B)$. 
Thus, we can rewrite the objective function Eq.~(\ref{eqn:loss1}) as 
\begin{equation}
    \max_{P\in\mathcal{P}} \left( \sum_{i,j}P_{ij}U_{ij} + \lambda tr(P^TD_A^TPD_B) \right).
    \label{eqn:loss2}
\end{equation}

The objective function Eq.~(\ref{eqn:loss2}) is defined on discrete space and is known as an NP problem.
Many recent methods~\cite{gao_deep_2021,wang_functional_2020,jaggi2015global} first relax the feasible field $\mathcal{P}$ to a continuous field $\mathcal{P'}$,
\begin{equation}
    \mathcal{P'} \triangleq \big\{P\in [0,1]^{n\times m};~P\mathbf{1_m}=\mathbf{1_n},~P^T\mathbf{1_n} \le \mathbf{1_m} \big\},
\end{equation}
then adopt gradient descend based methods to solve it. 

However, the second term in Eq.~(\ref{eqn:loss2}) is quadratic and its Hessian matrix is indefinite.
Hence, this objective function is nonconcave and has many local maximums. 
Existing solvers often require long computing time to converge.

\subsection{Our Model}
\label{sec:ourModel}

\paragraph{Symmetric Edge Matrices.}
We first analyze the properties of edge attribute matrices $D_A$ and $D_B$. 
Recent graph matching models construct them using pairwise structure information (\eg, Euclidean distance~\cite{wang_functional_2020}, adjacency~\cite{gao_deep_2021,wang2019learningIPCA}, Mahalanobis distance~\cite{gao_deep_2021}, and inner-product distance~\cite{sarlin2020superglue,sun2021loftr}) between feature points. In most of these models, $D_A$ and $D_B$ are symmetric matrices. 
However, they are usually not \emph{positive semi-definite}. 

If we can make the edge weight matrix $D$ \emph{positive semi-definite} 
(Sec.~\ref{sec:attr} will discuss the conversion of given $D_A$ and $D_B$ into positive semi-definite matrices.), then it can be decomposed as $D = HH^T$. 
Based on this, we can design a fast linear graph matching model. 


\paragraph{Linear Matching Model.} 
If $D_A$ and $D_B$ are positive semi-definite, then we can decompose them as $D_A = H_AH_A^T$ and $D_B = H_BH_B^T$. 
The quadratic term in Eq.~(\ref{eqn:loss2}) can be written as
\begin{equation}
    \begin{split}
        & tr(P^TD_A^TPD_B)\\
        = & tr(P^TD_APD_B) = tr(P^TH_AH_A^TPH_BH_B^T) \\
        = & tr(H_B^TP^TH_AH_A^TPH_B) = tr((H_A^TPH_B)^T(H_A^TPH_B))\\
        = & ||H_A^TPH_B||_F^2 = \sum_{i,j} (|H_A^TPH_B|_{ij})^2.\\
    \end{split}
    \label{eqn:sum-of-square}
\end{equation}
This term Eq.~(\ref{eqn:sum-of-square}) is quadratic and is in the form of the sum of squares (\ie, L2 norm). This converts the matching model Eq.~(\ref{eqn:loss2}) from a \emph{nonconvex-nonconcave} function to a \emph{convex} function.  
Nevertheless, maximizing a convex function still results in multiple local maximum and is difficult to solve. Our idea is to change this sum-of-squares form into a sum of absolute values (\ie, L1 norm). 
This modification has two main benefits:
(1) Compared with L2 norm, optimizing L1 norm promotes sparsity and is more robust against outliers. 
L1 norm has been used in PCA~\cite{kwak2008principal,ding2006r} and kernel discriminant analysis~\cite{zheng2013l1}, and demonstrated its better robustness when outliers exist.
(2) Numerically, combining with the Entropy regularization, the objective function can be made concave. Maximizing a concave objective function is much easier and it quickly converges to a global maximum.

Thus, we propose the \textbf{linear objective function} for graph matching, 
\begin{equation}
    \max_{P\in\mathcal{P'}} \left( \sum_{i,j}P_{ij}U_{ij} + \lambda\sum_{i,j} |H_A^TPH_B|_{ij} \right).
    \label{eqn:loss3}
\end{equation}

Inspired by the recent successful usage of Sinkhorn algorithm~\cite{cuturi_sinkhorn_2013} in Optimal Transport problem, in Eq.~(\ref{eqn:loss3}), we add an \emph{Entropy Regularization} $h(P) = -\sum_{i,j}P_{ij}\log P_{ij}$. Hence, finally, we have 
\begin{equation}
    \max_{P\in\mathcal{P'}} \left( \sum_{i,j}P_{ij}U_{ij} + \lambda\sum_{i,j} |H_A^TPH_B|_{ij} + \epsilon h(P) \right),\\
    \label{eqn:loss4}
\end{equation}
where $\epsilon > 0$ balances the weight of $h(P)$. 

This objective function Eq.~(\ref{eqn:loss4}) has negative semi-definite Hessian matrix $\mathcal{H}$, whose diagonal elements $\mathcal{H}_{ij, ij} = -\epsilon / P_{ij}$ and non-diagonal elements are all zeros.
Because $P_{ij} \ge 0$, the $\mathcal{H}$ is negative semi-definite.
Thus, this objective function Eq.~(\ref{eqn:loss4}) is concave and has a global maximum. 
And it can be solved efficiently using Lagrange multipliers (optimization details elaborated in Sec.~\ref{sec:solver}).


\subsection{Graph Attribute Construction}
\label{sec:attr}

When formulating the feature matching problem on a graph, 
both \emph{node similarity} (that aligns keypoints with similar descriptors) and \emph{structure similarity} (that matches relative positional or contextual correlation between keypoints) should be considered. These similarities are encoded in the matching of nodes and edges of the graph using the \emph{node similarity term} and \emph{edge similarity term}, respectively.

\paragraph{Node Similarity.}
Recent matching models first use neural networks to extract keypoints and their local feature descriptors, then compute the node similarity by a direct inner-product or a learnable metric~\cite{sarlin2020superglue,wang2019learningIPCA,gao_deep_2021,sun2021loftr}.
We chose Gao~\etal~\cite{gao_deep_2021}'s model as our baseline as it has the state-of-the-art accuracy.
Our main design is to speedup the more expensive edge similarity term. Hence, for a fair comparison, 
on the node similarity term,  
we followed the same setting of \cite{gao_deep_2021} and used VGG16 to extract descriptors and get the same node similarity matrix.

Let $\psi(V_A) \in \mathbb{R}^{n \times d}$ and $\psi(V_B)  \in \mathbb{R}^{m \times d}$ represent the feature maps of node $V_A$ and $V_B$.
The $i$th row of the feature map is the descriptor of the $i$th node. 
The node similarity matrix $U_{ij}$ is computed like the   Mahalanobis distance,
\begin{equation}
    U_{ij} = \psi(V_A) \Sigma_U \psi(V_B)^T,
\end{equation}
where $\Sigma_U \in \mathbb{R}^{n\times m}$ is a learnable matrix.
After the training stage, $\Sigma_U$ is fixed during the  inference stage.

\paragraph{Edge Similarity.}

Edge attribute matrices $D_A$ and $D_B$ are designed to characterize the structure of the matching model.
As we discussed in Sec.~\ref{sec:ourModel}, we want to develop a scheme to convert any given symmetric edge attribute matrix $D$ to a positive semi-definite matrix $\hat{D}$.

In graph matching models, because edges are usually undirected and unordered, most edge attributes that represent the inter-node relationship in graphs, such as Euclidean distance~\cite{wang_functional_2020}, adjacency~\cite{gao_deep_2021,wang2019learningIPCA}, Mahalanobis distance~\cite{gao_deep_2021}, and inner-product distance~\cite{sarlin2020superglue,sun2021loftr}, inherently form symmetric attribute matrices.  
Let $D$ be an edge attribute matrix, where $D_{ij}$ represents the edge attribute between node $v_i$ and $v_j$. 
$D$ usually has the following form: 
(1) When $i\neq j$, we have $D_{ij} = D_{ji}$.
(2) When $i = j$, a diagonal entry $D_{ii}$ is undefined and is set to $0$ in most existing models.

Given the two edge attributes $D_A$ and $D_B$, we can convert them to positive semi-definite matrices $\hat{D_A}$ and $\hat{D_B}$ by just modifying the diagonal entries, 
\begin{equation}
        d_x = \max \big\{R_i = \sum_{k\neq i}|(D_x)_{ik}|, i\in\{1, \cdots, n\}\big\}, d_{max} = \max \{d_A, d_B\},
\end{equation}
where $x=\{A, B\}$, $d_A$ and $d_B$ are the maximums of the sum of the absolute entries in each rows, and $d_{max}$ is the maximum of $d_A$ and $d_B$.
We use $d_{max}$ to modify both edge attributes to make sure they have the same diagonal entries,
\begin{equation}
    (\hat{D}_x)_{ij} = \begin{cases}
         d_{max} & i=j \\
         (D_x)_{ij} & i\neq j
    \end{cases},~x=\{A,B\}.
    \label{eqn:D_hat}
\end{equation}
We use $\hat{D}_A$ and $\hat{D}_B$ as the new edge attribute matrices to replace the original $D_A$ and $D_B$ in graph matching computation. 

$\hat{D}_A$ and $\hat{D}_B$ have two important properties:
    (1) \emph{Similar structural constraint}. The non-diagonal entries of $\hat{D}_x$ and the original one $D_x$ are the same, which characterize the structure and the edge attributes of the graph.
    Recall the structure constraints in the objective function Eq.~(\ref{eqn:loss1}), $||\hat{D}_A - P\hat{D}_BP^T||_F^2$.
    By definition, $P$ and $P^T$ can be treated as permutation matrices that switch the rows and columns of $\hat{D}_B$. 
    Hence, $P\hat{D}_BP^T$ has the same diagonal entries as $D_B$. 
    Then, since $\hat{D}_A$ and $\hat{D}_B$ have the same diagonal entries $d_{max}$, optimizing $||D_A - PD_BP^T||_F^2$ and $||\hat{D}_A - P\hat{D}_BP^T||_F^2$ is equivalent. 
    Modifying $D$ to $\hat{D}$ does not affect the optimal $P$.
    (2) \emph{Positive Semi-definiteness}. 
    They are both positive semi-definite and can be decomposed to construct the linear matching model. 
    We prove the positive semi-definiteness property by Gershgorin circle theorem~\cite{varga2010gervsgorin}, every eigenvalue of a matrix $M$ lies within at least one of the Gershgorin discs $r(\hat{D}_{ii}, R_i)$.

\subsection{CLAP Solver}
\label{sec:solver}


With positive semi-definite edge attribute matrices and their decomposition, our matching model Eq.~(\ref{eqn:loss4}) is concave and has a global maximum. 
This model can be solved efficiently using a Lagrangian multiplier method. 

Let $\mathcal{L}(P, \mu_1, \mu_2)$ be the Lagrangian of Eq.~(\ref{eqn:loss4}) with dual variables $\mu_1 \in \mathbb{R}^n, \mu_2 \in \mathbb{R}^m$,
\begin{equation}
    \begin{split}
        \mathcal{L}(P, \mu_1, \mu_2) =& \sum_{i,j} P_{ij} U_{ij} + \lambda \sum_{i,j} |H_A^TPH_B|_{ij} - \epsilon \sum_{i,j} P_{ij}log P_{ij} \\
        + & \mu_1^T(P\mathbf{1_m}-\mathbf{1_n}) + \mu_2^T(P^T\mathbf{1_n} -\mathbf{1_m}).
    \end{split}
    \label{eqn:la}
\end{equation}
For any couple $(i,j)$, let the first derivative of Eq.~(\ref{eqn:la}) be zero, we have
\begin{equation}
    \begin{split}
        0 =  U_{ij} - \epsilon - \epsilon logP_{ij} + (\mu_1)_i + (\mu_2)_j + \lambda \sum_{k=1}^{k_1}\sum_{l=1}^{k_2}\delta\left( (H_A^TPH_B)_{kl}\right) (H_A)_{ik}(H_B)_{jl},
    \end{split}
    \label{eqn:la_first}
\end{equation}

where $H_A\in \mathbb{R}^{n\times k_1}$, $H_B\in\mathbb{R}^{m\times k_2}$, and $\delta(\cdot) = \{-1, 1\}$ is the sign function.
The solution for Eq.~(\ref{eqn:la_first}) is,
\begin{equation}
    \begin{split}
        & P_{ij} = \exp\left( \frac{(\mu_1)_i}{\epsilon}-\frac{1}{2} \right) \exp \left(\frac{m_{ij}}{\epsilon}\right) \exp \left(\frac{(\mu_2)_j}{\epsilon}-\frac{1}{2} \right),P\mathbf{1_m} = \mathbf{1_n},P^T\mathbf{1_n} = \mathbf{1_m},
    \end{split}
    \label{eqn:sol-p}
\end{equation}
where $m_{ij} = U_{ij} + \lambda \sum_{k,l}\delta((H_A^TPH_B)_{kl})(H_A)_{ik}(H_B)_{jl}$. 
Note that since $\delta$ is a sign function, $m_{ij}$ is a scalar that does not contain the variable $P_{ij}$.
We substitute the $P_{ij}$ in Eq.~(\ref{eqn:sol-p}) back to the constraints of feasible field $\mathcal{P'}$.
By Cuturi's algorithm~\cite{cuturi_sinkhorn_2013}, $\mathcal{L}(P, \mu_1, \mu_2)$ has maximum point and can be computed with Sinkhorn's fixed point iteration.
For numerical stability, we follow Cuturi and Peyre~\cite{peyre2019computational} to solve the Eq.~(\ref{eqn:sol-p}) in log-domain.
Finally, we follow~\cite{gao_deep_2021,yu2019learningCIEH} to use the Hungarian algorithm for a discrete solution. 

\begin{table}[t]
  \centering
  \caption{Comparing the baseline and the revised CLIP model on synthetic image pairs. Three types of edge attributes (\ie, learning-based, adjacency matrix, and edge length distance) are tested. The baseline model is qc-DGM~\cite{gao_deep_2021}. 
  }
  \begin{tabular}{|@{}l|ccc@{}|}
    \hline
    Method         &Acc. (\%) & Time (ms) & FPS\\
    \hline
    Learning : qc-DGM    & 42.6 & 228.4 & 4.4 \\
    Learning : Ours      & \textbf{44.3} & \textbf{159.8} & \textbf{6.3} \\
    \hline
    Adjacency : qc-DGM     & 32.8 & 21.6 & 46.3 \\
    Adjacency : Ours       & \textbf{79.2} & \textbf{9.0} &\textbf{111.1} \\
    \hline
    Length : qc-DGM       & 63.8 & 22.8 & 43.9 \\
    Length : Ours         & \textbf{98.1} & \textbf{8.7} & \textbf{114.9} \\
    \hline
  \end{tabular}
  \label{tab:synth-result_pairwise}
\end{table}

\begin{table}[t]
  \centering
  \caption{Comparisons on synthetic image pairs. 
  }
  \begin{tabular}{|c|@{}l|ccc@{}|}
    \hline
    & Method         &Acc. (\%) & Time (ms) & FPS\\
    \hline
    \multirow{3}{*}{Unary} & IPCA~\cite{wang2019learningIPCA} & 23.7 & 207.1 & 4.83\\
        & PCA~\cite{wang2020combinatorialPCA} & 30.4 & 204.2 & 4.90 \\
        & CIE-H~\cite{yu2019learningCIEH} & 37.3 & 202.5 & 4.93 \\
    \hline
    \multirow{2}{*}{Pairwise} & qc-DGM~\cite{gao_deep_2021}    & 42.6 & 228.4 & 4.4 \\
    & Ours                  & \textbf{44.3} & \textbf{159.8} & \textbf{6.3} \\
    \hline
  \end{tabular}
  
  \label{tab:synth-unary-pairwise}
\end{table}

\begin{figure*}[t]
    \begin{center}
    \begin{tabular}{ccc}
         \includegraphics[width=0.3\linewidth]{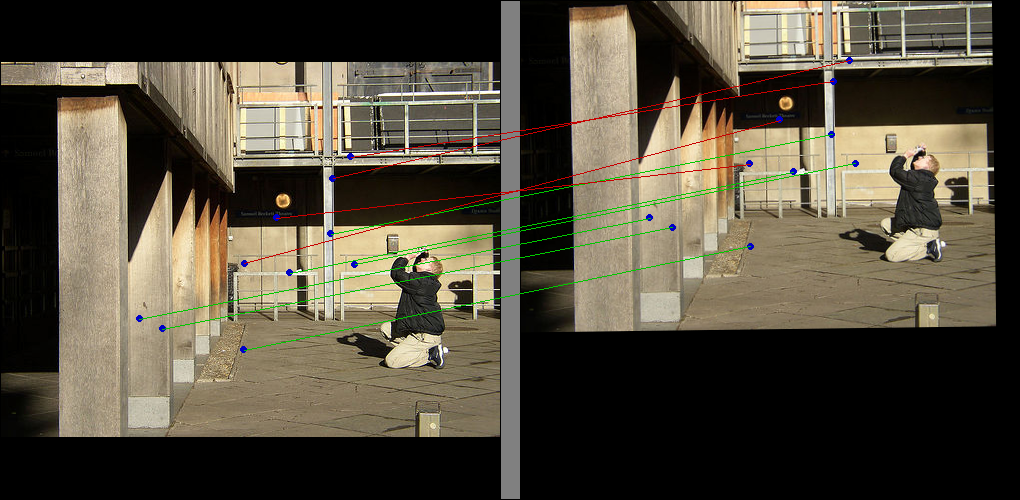} &
         \includegraphics[width=0.3\linewidth]{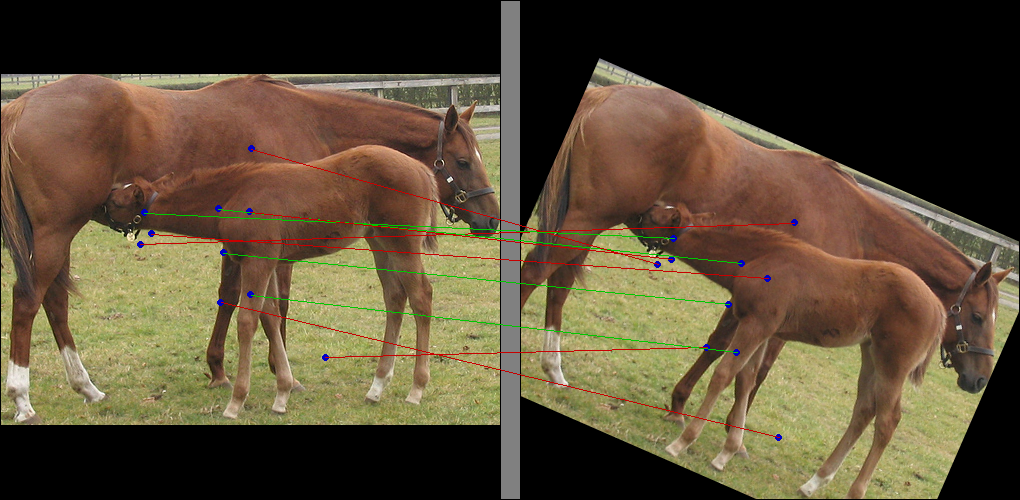} &
         \includegraphics[width=0.3\linewidth]{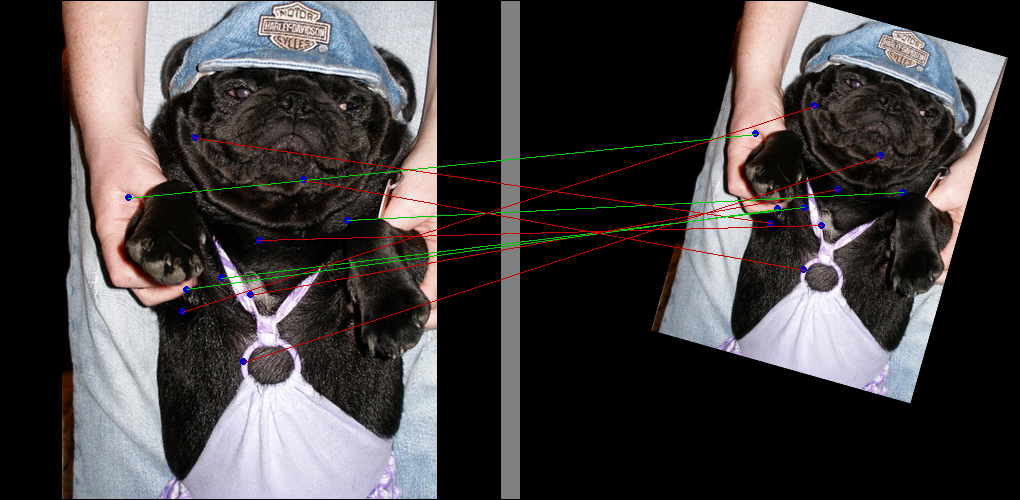}\\
         qc-DGM (Learning) Acc. 60\%  & qc-DGM (Adj) Acc. 40\% & qc-DGM (Len) Acc. 40\%\\
         \includegraphics[width=0.3\linewidth]{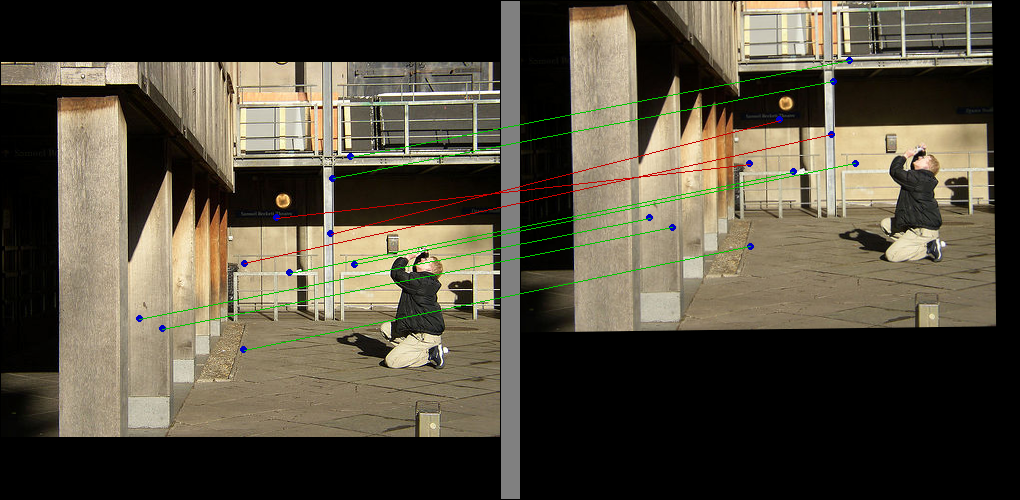}&
         \includegraphics[width=0.3\linewidth]{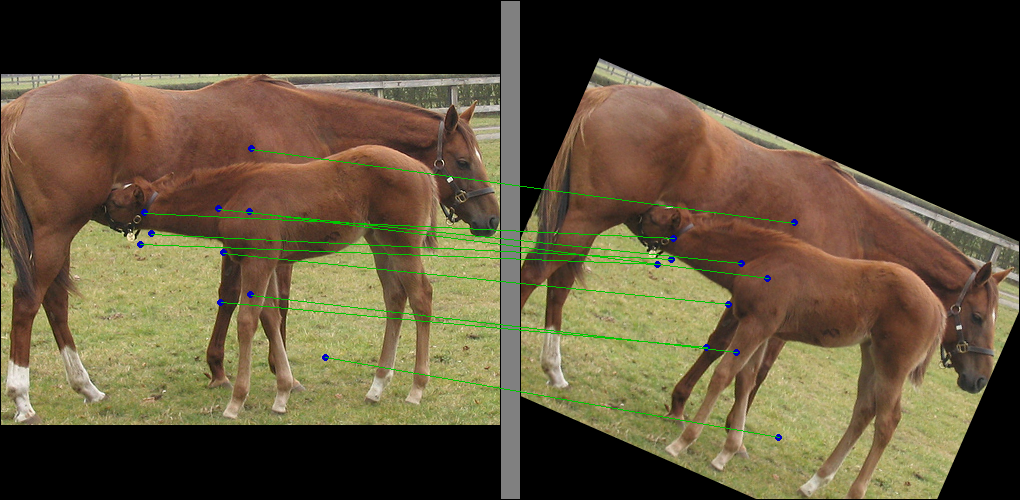}&
         \includegraphics[width=0.3\linewidth]{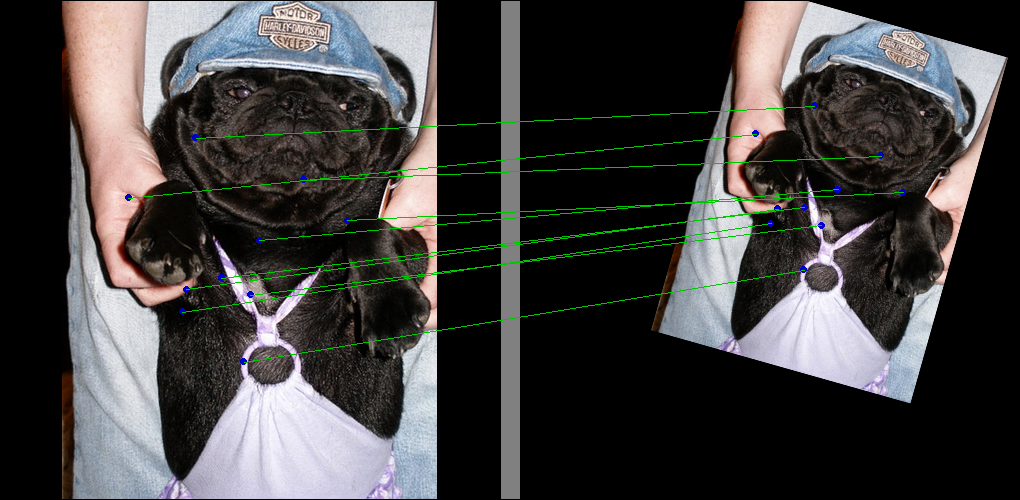}\\
         Ours (Learning) Acc. 70\% & Ours (Adj) Acc. 100\% & Ours (Len) Acc. 100\%\\
     \end{tabular}
    \end{center}
    \caption{Qualitative comparisons on synthetic transformed images, where \textcolor{green}{green}/\textcolor{red}{red} lines indicate correct/wrong matchings, respectively. `Adj', `Len', and `Learning' represent the three types of edge attributes adopted in qc-DGM and our models, respectively.}
    \label{fig:qual-synth}
\end{figure*}

\section{Experiments}

We conducted experiments to evaluate the proposed CLAP model. 
We evaluated our matching model's performance under various edge attributes on image pairs that differ by randomly synthesized geometric transformations.
We also compared our model with the SOTA graph matching algorithms.

\paragraph{Evaluation Metrics.} 
We compared \emph{matching accuracy} and \emph{matching efficiency}. 
A widely adopted \emph{matching accuracy} metric is the accuracy score \emph{Acc}. 
Given an assignment matrix $P$ of $n$ nodes, \emph{Acc} is defined as
    $Acc = \frac{1}{n} \sum_{ij} (P^* \circ P)_{ij},$
where $P^* \in \{0, 1\}^{n \times n}$ is the groundtruth correspondence, and operator $\circ$ is the element-wise production. 
To evaluate \emph{matching efficiency}, we measured the average per graph matching time: \emph{Time} and \emph{FPS} (frame-per-second). 

We compared our model with state-of-the-art matching methods, including \textbf{unary matching} models: PCA~\cite{wang2020combinatorialPCA}, IPCA~\cite{wang2019learningIPCA}, and CIE-H~\cite{yu2019learningCIEH}, and \textbf{pairwise matching} models: GMN~\cite{zanfir2018deepGMN}, NGM~\cite{wang2021neuralNGM}, HNN-HM~\cite{liao2021hypergraphHNN-HM}, and qc-DGM~\cite{gao_deep_2021}. 
Among these methods, since no codes nor pretrained models of HNN-HM~\cite{liao2021hypergraphHNN-HM}, LCSGM~\cite{LCSGMwang2020learning}, and GLMNet~\cite{jiang2019glmnet} is available, we use the evaluation scores from their papers directly.  
For other methods, we ran their pretrained models.
We set $\lambda=0.1$ and $\epsilon=1$.
All the experiments were done on an Intel Xeon(R) CPU E5-2630 with one NVIDIA GeForce 1080Ti graphic card.
Source code: \url{https://github.com/xmlyqing00/clap}.

\subsection{Results on Synthetic Transformations}

\paragraph{Dataset Construction.}
We synthesized $1,000$ pairs of images to evaluate the robustness of graph matching models under random affine transformations. 
The backgrounds are randomly selected from Pascal VOC dataset~\cite{everingham2010pascal}.
For each pair of images $(I_A, I_B)$, image $I_A$ contains $10$ nodes with random 2D-coordinates.
Then we synthesized an affine transformation with randomly generated scaling (range: $[0.5, 1)$), rotation (range: $[-\pi, \pi)$), and translation (range: $[-w/4, w/4)$ on x-axis and $[-h/4, h/4)$ on y-axis).
This affine transformation was applied on $I_A$ (and its node positions) to get $I_B$ (and the new node positions). 

\paragraph{Different Edge Attribute Constructions.}
Our proposed CLAP model is compatible with various edge attributes that are used in existing pairwise matching methods.
We tested on three types of commonly used edge attributes: (1) learning based edge descriptors~\cite{gao_deep_2021}, (2) 0/1 adjacency matrix
~\cite{zhou2012factorizedFGM,gao_deep_2021}, and (3) edge length structure~\cite{wang_functional_2020}. We denote them as ``Learning'', ``Adjacency'', and ``Length'' attributes in the following.

As shown in Tab.~\ref{tab:synth-result_pairwise}, our linear model not only preserves but actually surpasses the quadratic baseline model qc-DGM~\cite{gao_deep_2021} in matching accuracy, on all the three edge attribute settings. 
And our model is also significantly faster. 

One main reason that our linear model often leads to more accurate results in this experiment is that the quadratic baseline model is nonconcave and has multiple local maximums, especially when the image transformation is significant. 
In contrast, our linear model has a global maximum and is easy to be solved. 
This experiment demonstrates that our model (using L1 norm) is better than the baseline (using L2 norm).
Note that here the learning-based edge attributes were constructed using the pre-trained descriptors learned from the large benchmark dataset Pascal VOC. 
Their matching accuracy is low, showing that such learned descriptors don't generalize well for these synthesized transformations (probably much more significant than those in the benchmark).
Qualitative comparisons are shown in Fig.~\ref{fig:qual-synth}.

\begin{table*}[t]
  \centering
  \caption{Quantitative comparisons on Pascal VOC benchmark. We used the pretrained models that were trained on the same benchmark. 
  Our baseline is qc-DGM~\cite{gao_deep_2021}. $^+$ indicates an additional post-processing.
  The accuracy scores are in percentage. \textcolor{red}{Red} numbers indicate the best performance, and \textcolor{blue}{Blue} numbers indicate the second best.}
  \resizebox{\columnwidth}{!}{%
  \begin{tabular}{|l|ccccccccccc@{}|}
    \hline
    Method & GMN & PCA & NGM & IPCA & GLM & HNN-HM & LCSGM & CIE-H & qc-DGM & qc-DGM$^+$  & Ours\\
    \hline
    areo    & 40.8 & 49.0 & 52.5 & 54.0 & 52.0 & 39.6 & 46.9 & 51.7 & 49.3 & 49.8 & 49.2\\
    bike    & 58.0 & 60.8 & 62.0 & 64.9 & 67.3 & 55.7 & 58.0 & 67.6 & 65.6 & 66.9 & 65.7\\
    bird    & 59.8 & 65.1 & 62.5 & 64.8 & 63.2 & 60.7 & 63.6 & 70.0 & 60.8 & 62.0 & 61.1\\
    boat    & 50.5 & 58.2 & 59.2 & 61.5 & 57.4 & 76.4 & 69.9 & 61.1 & 56.4 & 56.9 & 56.2\\
    bottle  & 78.5 & 77.3 & 78.4 & 78.9 & 80.3 & 87.3 & 87.8 & 82.4 & 82.5 & 82.6 & 81.9\\
    bus     & 69.5 & 73.9 & 77.1 & 73.8 & 74.6 & 86.2 & 79.8 & 76.0 & 78.9 & 78.9 & 78.8\\
    car     & 65.9 & 65.7 & 73.8 & 71.7 & 70.0 & 77.6 & 71.8 & 70.6 & 71.9 & 72.3 & 72.1\\
    cat     & 64.7 & 68.4 & 68.1 & 70.9 & 72.6 & 54.2 & 60.3 & 71.7 & 71.3 & 71.6 & 71.5\\
    chair   & 40.3 & 42.9 & 43.8 & 46.6 & 38.9 & 50.0 & 44.8 & 43.5 & 41.7 & 42.8 & 42.1\\
    cow     & 61.8 & 63.9 & 66.6 & 66.2 & 66.3 & 60.7 & 64.3 & 70.5 & 67.7 & 67.9 & 67.3\\
    table   & 66.8 & 45.2 & 48.5 & 40.3 & 77.3 & 78.8 & 79.4 & 63.5 & 73.4 & 77.5 & 77.5\\
    dog     & 62.3 & 68.2 & 63.5 & 68.3 & 65.7 & 51.2 & 57.5 & 71.3 & 64.4 & 65.3 & 64.0\\
    horse   & 62.4 & 66.6 & 65.3 & 67.1 & 67.9 & 55.8 & 64.4 & 70.9 & 70.7 & 71.5 & 69.7\\
    mbike   & 58.9 & 61.5 & 63.0 & 65.1 & 64.2 & 60.2 & 57.6 & 66.8 & 65.8 & 66.3 & 65.9\\
    person  & 37.2 & 44.2 & 47.6 & 49.3 & 44.8 & 52.5 & 52.4 & 47.3 & 48.2 & 48.8 & 47.4\\
    plant   & 79.1 & 83.0 & 83.0 & 85.7 & 86.3 & 96.5 & 96.1 & 85.7 & 91.5 & 93.0 & 92.5\\
    sheep   & 66.8 & 66.7 & 67.3 & 68.9 & 69.0 & 58.7 & 62.9 & 69.0 & 68.4 & 69.5 & 69.0\\
    sofa    & 49.9 & 57.4 & 62.3 & 60.0 & 61.9 & 68.4 & 65.8 & 61.3 & 66.1 & 65.7 & 63.8\\
    train   & 85.5 & 78.3 & 80.3 & 82.4 & 79.3 & 96.2 & 94.4 & 83.5 & 88.1 & 88.1 & 88.0\\
    tv      & 91.0 & 89.1 & 90.0 & 88.6 & 91.3 & 92.8 & 92.0 & 89.7 & 92.0 & 92.1 & 91.8\\
    \hline
    Mean (\%)   & 62.5 & 64.3 & 65.8 & 66.5 & 67.5 & 68.0 & 68.5 & 68.7 & 68.7 & \textcolor{red}{69.5} & \textcolor{blue}{68.8}\\
    Time (ms)   & 89.4 & 89.2 & 105.2 & 94.9 & - & - & - & \textcolor{blue}{83.3} & 116.4 & 145.5 & \textcolor{red}{73.7} \\
    FPS         & 11.19 & 11.21 & 9.51 & 10.54 & - & - & - & \textcolor{blue}{12.00} & 8.59 & 6.87 & \textcolor{red}{13.57}\\
    \hline
  \end{tabular}
  }
  
  \label{tab:pascal}
\end{table*}

  

\paragraph{Structure Info Encoding: Unary vs Pairwise Matching.}
We also compared the matching computed by unary models and pairwise models in Tab.~\ref{tab:synth-unary-pairwise}. 
Three representative unary models, IPCA~\cite{wang2019learningIPCA}, PCA~\cite{wang2020combinatorialPCA} and CIE-H~\cite{yu2019learningCIEH}, and two pairwise models, qc-DGM~\cite{gao_deep_2021} and our CLAP, are compared. 
Unary matching models learn to encode structure information by aggregating node descriptors. 
For all these five methods, we used node and edge attributes from the models pretrained on the Pascal VOC benchmark~\cite{everingham2010pascal}.
Tab.~\ref{tab:synth-unary-pairwise} shows the matching results on synthesized data. 
In this synthetic experiment, due to the significant rotation components involved, these learned/aggregated descriptors tend to be not discriminating enough to support reliable unary matching. 
Pairwise matching, in contrast, uses the explicit structure constraint, and turns out to be more reliable. 
In terms of efficiency, 
although unary matching methods have simpler matching model to solve, they need to conduct (expensive) aggregation to encode structure information from node descriptors. This makes their overlap matching speed slower than our CLAP model.

\subsection{Results of Pascal VOC Keypoints}

Pascal VOC dataset~\cite{everingham2010pascal} with Berkeley annotations of keypoints has 20 classes of instance images with keypoints.
The training set includes 7,020 annotated images and the testing set includes 1,682 images.
Our baseline qc-DGM model~\cite{gao_deep_2021} has two settings: with and without a post-processing refinement, respectively denoted as qc-DGM and qc-DGM$^+$. 
Our baseline is based on qc-DGM without post-processing. 
For fair comparisons, we kept all the feature extraction and aggregation unchanged with the pretrained weights, and just replaced the KB-QAP objective function and its solver with our proposed algorithm.

We report the average runtime and accuracies in Tab.~\ref{tab:pascal}.
Unary matching methods, such as LCSGM~\cite{LCSGMwang2020learning} and CIE-H~\cite{yu2019learningCIEH}, are highly reliant on the node attribute learning.
They have to use extra embedding network to enhance the local features. Therefore, although without pairwise constraints, their matching is faster. Their feature learning component is slower. 
Lawler's QAP methods GMN~\cite{zanfir2018deepGMN} and NGM~\cite{wang2021neuralNGM} have lower accuracy scores. 
Our CLAP model formulate the graph matching by L1 norm that is linear to solve. Therefore, it is both accurate and efficient. 

Compared with the baseline qc-DGM, the proposed model achieves matching accuracy of $68.8\%$ (baseline $68.7\%$) and $73.7$ms computation time (baseline 116.4ms).
Our model achieves similar accuracy but significantly improves the runtime efficiency. 
Note that here our model does not include a post-processing.
Although qc-DGM$^+$, with post-processing refinement added to the baseline, achieves slightly better matching score, its computation speed is much slower. 
The results show that: (1) our positive semi-definite edge attribute matrices can successfully model structure information; (2) our CLAP model can greatly improve matching efficiency without losing accuracy.


\section{Conclusion}
This paper presents a new linear model for fast graph matching. 
We reformulated the pairwise graph matching as a concave maximization problem, which has a global maximum and can be solved efficiently.
Specifically, we converted the pairwise structure constraint of KB-QAP into an L1 norm linear model.
We showed that a common symmetric edge attribute matrix can be refined to become positive semi-definite to construct a linear structure constraint. Then, the problem can be solved using the Sinkhorn algorithm. 
Experiments showed that our method can achieve state-of-the-art performance and can greatly improve the computation speed of pairwise graph matching. 

\paragraph{Limitations.} 
Pointwise affinity depends on descriptor learning. 
When images undergo big global transformations or local deformation/transformation, descriptors can become unreliable and this could negatively impact matching accuracy. 
In the future, we will explore learning mechanisms 
to estimate confidence of local feature descriptors (\ie, pointwise affinity), 
and increase structure constraints when necessary. 
Such a refined adaptive matching model could potentially improve the matching robustness in these challenging scenarios.

\begin{credits}
\subsubsection{\ackname} 
This research was partially supported by NSF CBET-2115405 and the Texas A\&M University ASCEND Research Leadership Fellows Program. Part of the experiments were conducted using the computing resources provided by Texas A\&M High Performance Research Computing.
\end{credits}

%
%
%

\bibliographystyle{splncs04}
\bibliography{refs}

\begin{thebibliography}{10}
\providecommand{\url}[1]{\texttt{#1}}
\providecommand{\urlprefix}{URL }
\providecommand{\doi}[1]{https://doi.org/#1}

\bibitem{cuturi_sinkhorn_2013}
Cuturi, M.: Sinkhorn {Distances}: {Lightspeed} {Computation} of {Optimal}
  {Transport}. In: Burges, C.J.C., Bottou, L., Welling, M., Ghahramani, Z.,
  Weinberger, K.Q. (eds.) Advances in {Neural} {Information} {Processing}
  {Systems} 26 (2013)

\bibitem{detone2018superpoint}
DeTone, D., Malisiewicz, T., Rabinovich, A.: Superpoint: Self-supervised
  interest point detection and description. In: CVPR workshops. pp. 224--236
  (2018)

\bibitem{ding2006r}
Ding, C., Zhou, D., He, X., Zha, H.: R 1-pca: rotational invariant l 1-norm
  principal component analysis for robust subspace factorization. In: ICML
  (2006)

\bibitem{everingham2010pascal}
Everingham, M., Van~Gool, L., Williams, C.K., Winn, J., Zisserman, A.: The
  pascal visual object classes (voc) challenge. IJCV  \textbf{88}(2),  303--338
  (2010)

\bibitem{gao_deep_2021}
Gao, Q., Wang, F., Xue, N., Yu, J., Xia, G.: Deep graph matching under
  quadratic constraint. In: CVPR. pp. 5067--5074 (2021)

\bibitem{he2021learnable}
He, J., Huang, Z., Wang, N., Zhang, Z.: Learnable graph matching: Incorporating
  graph partitioning with deep feature learning for multiple object tracking.
  In: CVPR. pp. 5299--5309 (2021)

\bibitem{jaggi2015global}
Jaggi, M., Lacoste-Julien, S.: On the global linear convergence of frank-wolfe
  optimization variants. NeurIPS  \textbf{28} (2015)

\bibitem{jiang2019glmnet}
Jiang, B., Sun, P., Tang, J., Luo, B.: Glmnet: Graph learning-matching networks
  for feature matching. arXiv preprint arXiv:1911.07681  (2019)

\bibitem{johnson1979computers}
Johnson, D.S., Garey, M.R.: Computers and intractability: A guide to the theory
  of NP-completeness. WH Freeman (1979)

\bibitem{koopmans1957assignment}
Koopmans, T.C., Beckmann, M.: Assignment problems and the location of economic
  activities. Econometrica: journal of the Econometric Society  (1957)

\bibitem{kwak2008principal}
Kwak, N.: Principal component analysis based on l1-norm maximization. IEEE
  T-PAMI  \textbf{30}(9),  1672--1680 (2008)

\bibitem{lawler1963quadratic}
Lawler, E.L.: The quadratic assignment problem. Management science
  \textbf{9}(4) (1963)

\bibitem{liao2021hypergraphHNN-HM}
Liao, X., Xu, Y., Ling, H.: Hypergraph neural networks for hypergraph matching.
  In: ICCV. pp. 1266--1275 (2021)

\bibitem{lin2023graph}
Lin, Y., Yang, M., Yu, J., Hu, P., Zhang, C., Peng, X.: Graph matching with
  bi-level noisy correspondence. In: ICCV. pp. 23362--23371 (2023)

\bibitem{loiola2007survey}
Loiola, E.M., de~Abreu, N.M.M., Boaventura-Netto, P.O., Hahn, P., Querido, T.:
  A survey for the quadratic assignment problem. European journal of
  operational research  \textbf{176}(2),  657--690 (2007)

\bibitem{lu2016fast}
Lu, Y., Huang, K., Liu, C.L.: A fast projected fixed-point algorithm for large
  graph matching. Pattern Recognition  \textbf{60},  971--982 (2016)

\bibitem{peyre2019computational}
Peyr{\'e}, G., Cuturi, M., et~al.: Computational optimal transport: With
  applications to data science. Foundations and Trends in Machine Learning
  \textbf{11}(5-6) (2019)

\bibitem{puy2020flot}
Puy, G., Boulch, A., Marlet, R.: Flot: Scene flow on point clouds guided by
  optimal transport. In: ECCV. pp. 527--544. Springer (2020)

\bibitem{sarlin2020superglue}
Sarlin, P.E., DeTone, D., Malisiewicz, T., Rabinovich, A.: Superglue: Learning
  feature matching with graph neural networks. In: CVPR. pp. 4938--4947 (2020)

\bibitem{Shen19CVPR}
Shen, X., Wang, C., Li, X., Cheng, M., Yu, Z., Li, J., Wen, C., Cheng, M., He,
  Z.: Rf-net: An end-to-end image matching network based on receptive field.
  In: CVPR. pp. 8132--8140 (2019)

\bibitem{sun2021loftr}
Sun, J., Shen, Z., Wang, Y., Bao, H., Zhou, X.: Loftr: Detector-free local
  feature matching with transformers. In: CVPR. pp. 8922--8931 (2021)

\bibitem{umeyama1988eigendecomposition}
Umeyama, S.: An eigendecomposition approach to weighted graph matching
  problems. IEEE T-PAMI  \textbf{10}(5),  695--703 (1988)

\bibitem{varga2010gervsgorin}
Varga, R.S.: Ger{\v{s}}gorin and his circles. Springer Science \& Business
  Media (2010)

\bibitem{wang_functional_2020}
Wang, F.D., Xue, N., Zhang, Y., Xia, G.S., Pelillo, M.: A {Functional}
  {Representation} for {Graph} {Matching}. IEEE T-PAMI  \textbf{42}(11),
  2737--2754 (Nov 2020)

\bibitem{wang2023deep}
Wang, R., Guo, Z., Jiang, S., Yang, X., Yan, J.: Deep learning of partial graph
  matching via differentiable top-k. In: CVPR. pp. 6272--6281 (2023)

\bibitem{wang2019learningIPCA}
Wang, R., Yan, J., Yang, X.: Learning combinatorial embedding networks for deep
  graph matching. In: ICCV. pp. 3056--3065 (2019)

\bibitem{wang2020combinatorialPCA}
Wang, R., Yan, J., Yang, X.: Combinatorial learning of robust deep graph
  matching: an embedding based approach. IEEE T-PAMI  (2020)

\bibitem{wang2021neuralNGM}
Wang, R., Yan, J., Yang, X.: Neural graph matching network: Learning lawler’s
  quadratic assignment problem with extension to hypergraph and multiple-graph
  matching. IEEE T-PAMI  (2021)

\bibitem{LCSGMwang2020learning}
Wang, T., Liu, H., Li, Y., Jin, Y., Hou, X., Ling, H.: Learning combinatorial
  solver for graph matching. In: CVPR. pp. 7568--7577 (2020)

\bibitem{yi2016lift}
Yi, K.M., Trulls, E., Lepetit, V., Fua, P.: Lift: Learned invariant feature
  transform. In: ECCV. pp. 467--483. Springer (2016)

\bibitem{yu2019learningCIEH}
Yu, T., Wang, R., Yan, J., Li, B.: Learning deep graph matching with
  channel-independent embedding and hungarian attention. In: ICLR (2019)

\bibitem{zanfir2018deepGMN}
Zanfir, A., Sminchisescu, C.: Deep learning of graph matching. In: CVPR. pp.
  2684--2693 (2018)

\bibitem{zaslavskiy2008path}
Zaslavskiy, M., Bach, F., Vert, J.P.: A path following algorithm for the graph
  matching problem. IEEE T-PAMI  \textbf{31}(12),  2227--2242 (2008)

\bibitem{zhang2019kergm}
Zhang, Z., Xiang, Y., Wu, L., Xue, B., Nehorai, A.: Kergm: Kernelized graph
  matching. NeurIPS  \textbf{32},  3335--3346 (2019)

\bibitem{zheng2013l1}
Zheng, W., Lin, Z., Wang, H.: L1-norm kernel discriminant analysis via bayes
  error bound optimization for robust feature extraction. IEEE transactions on
  neural networks and learning systems  \textbf{25}(4),  793--805 (2013)

\bibitem{zhou2012factorizedFGM}
Zhou, F., De~la Torre, F.: Factorized graph matching. In: CVPR. IEEE (2012)

\end{thebibliography}

\end{document}